%% file: main.tex
\documentclass[conference]{IEEEtran}
\IEEEoverridecommandlockouts
\usepackage{cite}
\usepackage{amsmath,amssymb,amsfonts}
\usepackage{algorithmic}
\usepackage{graphicx}
\usepackage{textcomp}
\usepackage{xcolor}
\usepackage{hyperref}
\usepackage{subfigure}
\newcommand{\ts}{\textsuperscript}
\def\BibTeX{{\rm B\kern-.05em{\sc i\kern-.025em b}\kern-.08em
    T\kern-.1667em\lower.7ex\hbox{E}\kern-.125emX}}
\begin{document}

\title{Deep Spiking Convolutional Neural Network for Single Object Localization Based On Deep Continuous Local Learning\\

}

\author{\IEEEauthorblockN{Sami Barchid}
\IEEEauthorblockA{\textit{Univ. Lille, CNRS, Centrale Lille, UMR 9189 CRIStAL} \\
% \textit{Univ. Lille}\\
F-59000 Lille, France \\
sami.barchid@univ-lille.fr}
\and
\IEEEauthorblockN{José Mennesson}
\IEEEauthorblockA{
\textit{IMT Lille-Douai, Institut Mines-Télécom,} \\
\textit{Centre for Digital Systems}\\
\textit{Univ. Lille, CNRS, Centrale Lille, UMR 9189 CRIStAL}\\
F-59000 Lille, France \\
jose.mennesson@imt-lille-douai.fr}
\and
\IEEEauthorblockN{Chaabane Djéraba}
\IEEEauthorblockA{\textit{Univ. Lille, CNRS, Centrale Lille, UMR 9189 CRIStAL} \\
F-59000 Lille, France \\
chabane.djeraba@univ-lille.fr}
}

\maketitle

\begin{abstract}
With the advent of neuromorphic hardware, spiking neural networks can be a good energy-efficient alternative to artificial neural networks. However, the use of spiking neural networks to perform computer vision tasks remains limited, mainly focusing on simple tasks such as digit recognition. It remains hard to deal with more complex tasks (e.g. segmentation, object detection) due to the small number of works on deep spiking neural networks for these tasks. The objective of this paper is to make the first step towards modern computer vision with supervised spiking neural networks. We propose a deep convolutional spiking neural network for the localization of a single object in a grayscale image. We propose a network based on DECOLLE, a spiking model that enables local surrogate gradient-based learning. The encouraging results reported on Oxford-IIIT-Pet validates the exploitation of spiking neural networks with a supervised learning approach for more elaborate vision tasks in the future.
\end{abstract}


\input{1_introduction}
\input{3_related_works}
\input{2_background}
\input{4_methodology}
\input{5_experiments}
\input{6_conclusion}

% \begin{table}[htbp]
% \caption{Table Type Styles}
% \begin{center}
% \begin{tabular}{|c|c|c|c|}
% \hline
% \textbf{Table}&\multicolumn{3}{|c|}{\textbf{Table Column Head}} \\
% \cline{2-4} 
% \textbf{Head} & \textbf{\textit{Table column subhead}}& \textbf{\textit{Subhead}}& \textbf{\textit{Subhead}} \\
% \hline
% copy& More table copy$^{\mathrm{a}}$& &  \\
% \hline
% \multicolumn{4}{l}{$^{\mathrm{a}}$Sample of a Table footnote.}
% \end{tabular}
% \label{tab1}
% \end{center}
% \end{table}

%\section*{Acknowledgment}

%This work was partly supported by IRCICA USR 3380 (CNRS, Univ. Lille, F-59000 Lille, France).

\bibliographystyle{IEEEtran}
\bibliography{references}
\end{document}

%% file: 1_introduction.tex
\section{Introduction}
\label{sec:intro}

Computer vision has shown great progress with the advent of Artificial Neural Networks (ANN) and deep learning, which achieves state-of-the-art performance for most vision tasks \cite{survey-deep}. However, modern deep learning approaches require a high computational complexity, which leads to high energy consumption. Although many works focus on reducing the computational complexity of ANNs \cite{survey_DL_compression}, they are still energy-intensive compared to the biological brain, which only requires around two dozens of Watts for its full activity. Therefore, bio-inspired methods are good candidates to design low-power solutions and solve the problem of energy consumption.

Spiking Neural Networks (SNNs), often known as the third generation of neural networks \cite{maass1997networks}, are strongly inspired from biological neurons \cite{hodgkin1952quantitative}. They consist of spiking neurons, which communicate through discrete spatio-temporal events called "spikes". As opposed to ANNs, they are amenable to implementation in efficient and low-power neuromorphic hardware \cite{neuromorphic_STDP}. However, their performance is still behind ANNs. It can be explained by the non-differentiable nature of spikes, making it impossible to use the standard back-propagation algorithm \cite{BP_SNN}.

Concerning computer vision using SNNs, their applications in computer vision are still limited compared to ANNs. One of the reasons is that many research papers mainly focus on simple tasks like digit recognition \cite{falez2019multi}. Therefore, we argue that there is a lack of works aiming at more complex computer vision tasks with SNNs (e.g. object detection). 

Our objective in this paper is to validate the ability of SNNs trained with recent supervised methods \cite{tavanaei2019deep} to deal with more complex vision tasks. To do so, we propose a Deep Convolutional SNN (DCSNN) to perform object localization of one object in grayscale images, as a first step towards a fully functional object detection solution. Our proposed DCSNN is based on Deep Continuous Local Learning (DECOLLE) \cite{DECOLLE}, a spiking model based on supervised local synaptic plasticity rule. We report proof-of-concept results on Oxford-IIIT-Pet \cite{Oxford-IIIT-Pet} and discuss the perspective of this preliminary work.

The rest of the paper is organized as follows: Section \ref{sec:related_works} discusses the existing works on SNNs related to vision tasks and training methods. Section \ref{sec:background} formulates the preliminary notions on SNNs and briefly introduces DECOLLE. Section \ref{sec:methodology} describes our DCSNN approach based on DECOLLE. Section \ref{sec:experiments} reports our experimental proof-of-concept for object localization using DCSNN. Section \ref{sec:conclusion} concludes our work and discusses the perspectives for our approach.

%% file: 3_related_works.tex
\section{Related Works}
\label{sec:related_works}

Since spiking neurons are not differentiable, SNNs cannot directly benefit from the successful back-propagation algorithm \cite{BP_SNN}. Nevertheless, many other learning approaches exist and can vary according to the strategy employed. 

Some works \cite{Conversion_method} focus on converting a trained ANN into a functional SNN without losing too much performance. Consequently, the converted SNN can benefit from the best of both worlds, by achieving good performance of ANNs and being able to benefit from neuromorphic hardware's efficiency. This approach tends to achieve the best results in accuracy metrics \cite{tavanaei2019deep}. In addition, there already exist fully functional converted SNN that can deal with modern vision tasks \cite{conversion_sota}, such as Object Detection \cite{Spiking-YOLO}. However, the training of the original ANN remains energy-intensive and the SNN is not trained directly. This method can be seen as an optimization technique for ANNs rather than a learning rule for SNNs.

On the other hand, many works try to exploit bio-plausible learning rules, well-known in neuroscience, to train SNNs on computer vision tasks. One of the most used learning rules of this type is the Spike-Timing-Dependent Plasticity (STDP) \cite{STDP}. It updates the weights of the SNNs according to the relative spike times of pre- and post-synaptic neurons. This kind of approach is unsupervised and local, which means that they are suitable for neuromorphic hardware. However, building deep SNNs with bio-plausible learning remains an open problem, due to the need to design additional neural mechanisms \cite{falez2019multi}. Methods based on this strategy are usually limited to small networks and deal with simple tasks (e.g. digit recognition) because STDP-like approaches are still unable to effectively learn from complex natural images \cite{falez_thesis}.  However, recent works show great performance of SNNs trained with STDP when coupled to event-based cameras (e.g. DVS cameras). For instance, \cite{kirkland2020spikeseg} performs image segmentation from an event-based camera with a spike-based approach trained with STDP. 

As for training SNNs with supervised learning, many recent papers propose to adapt the back-propagation algorithm for SNNs. To do so, a popular direction is to back-propagate a surrogate gradient as a continuous relaxation of the real gradient \cite{zenke2018superspike, neftci2019surrogate,DECOLLE, lee2020enabling}. Although this strategy enables the use of supervised learning to train SNNs, there is still no works on vision tasks other than simple classification tasks. We argue that surrogate gradient approaches can be exploited to perform more complex vision tasks and, consequently, pave the way towards functional solutions. Among the existing works on surrogate gradient learning, DECOLLE \cite{DECOLLE} is a recent method that has many advantages: \textbf{1) }it enables local learning, making it implementable in neuromorphic hardware; \textbf{2)} it has low memory complexity compared to other methods; \textbf{3)} its simplicity makes it amenable to implementation on popular machine learning frameworks. For these reasons, we decide to use DECOLLE for our work in order to propose the first work towards modern vision tasks with supervised SNN. Sec. \ref{sec:background} briefly introduces DECOLLE approach.

%% file: 2_background.tex
\section{Background}
\label{sec:background}

\subsection{Spiking Neural Networks}
\label{subsec:spiking_neural_networks}
The main difference between artificial and spiking neurons is the representation of information. In a spiking neuron, information is represented with a series of short electrical impulses known as "spikes", as opposed to the numerical representation of artificial neurons. In addition, there are various neuron models to represent a spiking neuron.

\begin{figure}[ht]
\centering
\includegraphics[width=0.48\textwidth]{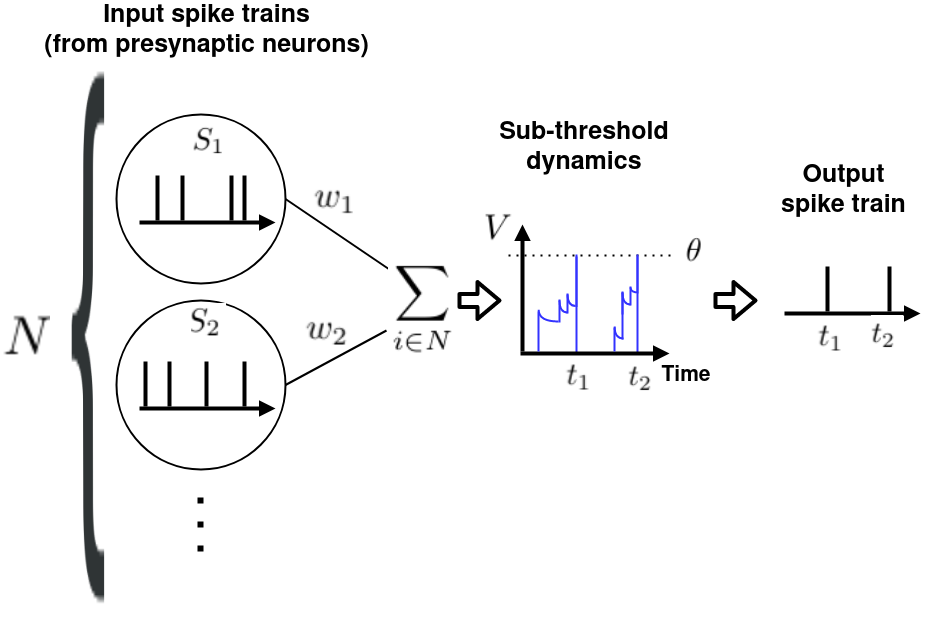}
\caption{The LIF neuron model dynamics (inspired from \cite{lee2020enabling}). The presynaptic spikes are weighted by the synaptic weights before being integrated in the neuron. The integrated spikes increases the leaky membrane potential until a defined threshold is reached. When the membrane potential exceeds the threshold, the neuron emits an output spike and the membrane potential is reset.}
\label{fig:LIF_Model}
\end{figure}

One of the most popular and simplest neuron models is the Leaky Integrate-and-Fire (LIF) neuron. It acts as a leaky integrator of input spikes to its internal membrane potential. When the membrane potential $V$ exceeds a defined firing threshold $\theta$, the neuron emits an output spike to its post-synaptic neurons, and its membrane potential is reset to a resting potential (here, set to 0). The model can be formulated as follows:
\begin{equation}
\begin{split}
\tau_{leak} \frac{dV}{dt} = - V + I(t)\\
V \leftarrow 0 \: when \: V \ge \theta
\end{split}
\end{equation},
where $\tau_{leak}$ is the membrane time constant (which describes the leaky dynamics) and $I(t)$ is the input current at the current time $t$. $I(t)$ is defined as the summation of the presynaptic spikes weighted by the related synaptic weights. The input current at a time $t$ is formulated as follows :

\begin{equation}
\begin{split}
I(t) = \sum_{i \in N}{( w_i \sum_{j}{S_i(t - t_j)} )}
\end{split}
\end{equation}
where $N$ is the set of presynaptic neurons and $w_i$ is the synaptic weight of the presynaptic neuron $i$. $S_i$ is the spike train of the presynaptic neuron $i$ where a spike occurs at time $t_j$ for the $j$\ts{th} spike. A spike event $S_i(t - t_j)$ can be described as a Kronecker delta function :
\begin{equation}
\begin{split}
S_i(t - t_j) = 
\begin{cases}
    1, &         \text{if } t = t_j\\
    0, &         \text{otherwise }
\end{cases}
\end{split}
\end{equation}

. Figure \ref{fig:LIF_Model} shows the dynamics of the formulated LIF model.

SNNs are trained by updating the synaptic weights between connected neurons, similarly to ANNs. An SNN can be structured in the same way as an ANN, with neurons organized into successive layers (e.g. convolutional layers).

Because developing SNNs directly on neuromorphic hardware can be a daunting task, SNNs are firstly simulated on GPU/CPU hardware. The simulation requires to discretize the continuous dynamics of spiking neurons into successive timesteps. For the simulation, we define $T$ timesteps of 1ms duration.

\subsection{Deep Continuous Local Learning}
  \label{subsec:DECOLLE}
DECOLLE \cite{DECOLLE} is a supervised approach based on surrogate gradient learning for SNNs. Since the synaptic weights are updated after each timestep, the DECOLLE rule is used for online learning.
 
 In DECOLLE, a readout layer initialized with fixed random weights is attached to each LIF layer of the network. At each timestep, a readout layer multiplies the neural activation of its related LIF layer with its fixed random weights. The output of a readout layer is a prediction that is used to solve the targeted task and compute a local error. Consequently, each layer minimizes its own local error function. LIF layers build on the features of the previous layers trained with their own local error function. The layers indirectly learn features that are useful for their successors.  
 
The advantage of this approach is the locality of the error optimization, which makes it usable on neuromorphic hardware. Another advantage is the low memory complexity required to simulate a DECOLLE network with popular auto-differentiation frameworks (e.g. PyTorch). 

%% file: 4_methodology.tex
\section{Methodology}
\label{sec:methodology}

\subsection{Problem Formulation}
\label{subsec:problem_formulation}
The targeted task is to perform object localization on strictly one object in a grayscale image. We define the problem as follows : given a grayscale image $\mathbf{I} \in \mathbb{R}^{H \times W}$, a bounding box of coordinates $\mathbf{B} = \{ x_{min}, y_{min}, x_{max}, y_{max} \}$\footnote{These coordinates correspond to the upper-left corner $(x_{min}, y_{min})$ and the lower-right corner $(x_{max},y_{max})$ of the predicted bounding box $\mathbf{B}$.} surrounding the object in the foreground is predicted. To do so, we propose $s(\mathbf{I})$, a DCSNN based on DECOLLE, which consists of $L$ convolutional LIF layers.

\begin{figure*}[ht]
\centering
\includegraphics[width=1.0\textwidth]{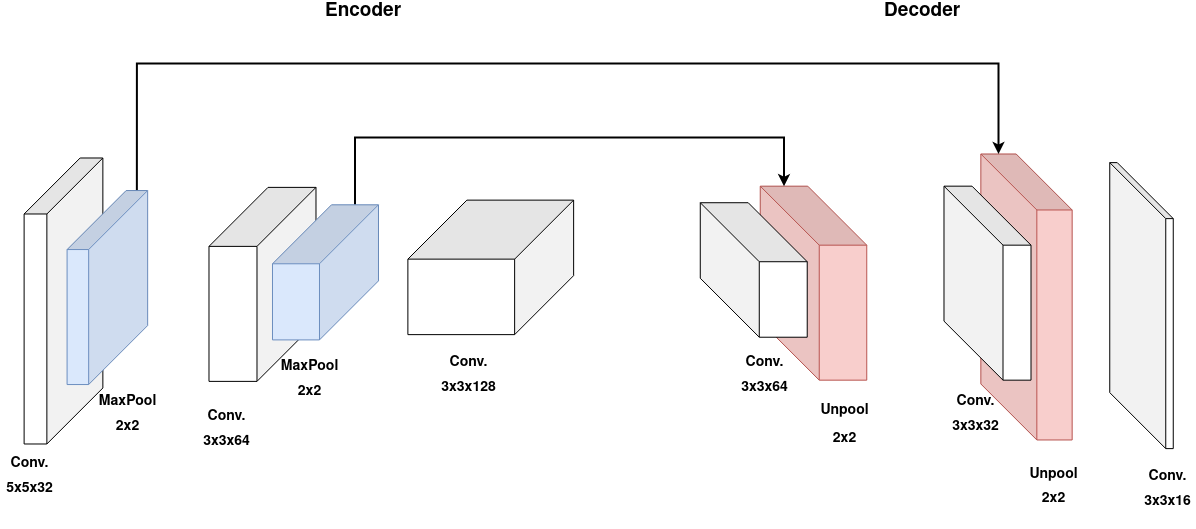}
\caption{Our proposed DCSNN architecture. White, blue and red cubes represent LIF convolutional layer, maxpool layer, and upsampling layer respectively.  Black arrows are residual connections between feature maps (i.e. addition operations). For simplification purposes, the readout linear layers (specific to DECOLLE \cite{DECOLLE} model) are not shown.}
\label{fig:FPN_DECOLLE}
\end{figure*}

\subsection{Readout Layers}
As stated in Sec. \ref{sec:background}, a readout layer is attached to each layer of spiking neurons in a DECOLLE model. According to our targeted problem formulated in Sec. \ref{subsec:problem_formulation}, the DCSNN $s(.)$ must predict the coordinates of a bounding box $\mathbf{B}$. Therefore, the readout layers are simply linear layers in order to predict the four needed coordinates of $\mathbf{B}$. In other words, our DCSNN $s(.)$ produces a series of $L$ bounding box predictions, i.e. $s(\mathbf{I}) = \{ \mathbf{B}_i\}^{L}_{1}$.

\subsection{Network Architecture}
Our proposed architecture is structured following the popular encoder-decoder paradigm used in common ANN models \cite{FPN}. Fig. \ref{fig:FPN_DECOLLE} shows the overall architecture of our model. It consists of an encoder composed of 3 convolutional LIF layers, which subsequently increase the semantic information and decrease the spatial information. Since the deep parts of an encoder tend to lose the spatial information of low-level layers \cite{FPN}, a decoder module is used to recover the spatial details. A decoder of 3 convolutional LIF layers is employed, where the output spikes from the encoder's layers are passed to the decoder through residual connections. The residual connections are essentially addition operations between two spiking feature maps.

\subsection{Static Image Neural Coding}
As an SNN uses spikes to transmit information, numerical values cannot be fed directly in a spike-based network. Pixels must beforehand be encoded into spike trains using a function known as \textit{neural coding} \cite{borst1999information}.

Among the existing neural coding strategies, we choose rate coding, which means pixel values are represented as a frequency of spikes during a defined time (i.e. a high pixel value results in a high frequency of spikes). Specifically, each numerical value of the pixels is converted using a Poisson process, which creates a spike train occurring at a target frequency with no spike-to-spike correlation. However, the main drawback of rate coding can be mentioned: a large number of spikes are needed to encode information, which introduces latency into the network \cite{falez_thesis}.

\subsection{Output Conversion Strategy}
\label{subsec:output_conversion_policy}
Since the objective of object localization is to predict the bounding box coordinates, the output spikes of an SNN must be converted back to numerical values. In DECOLLE, this problem is partially solved thanks to the readout layers, which return a numerical prediction at each of the $T$ timesteps. The remaining issue is the final prediction computed among the $T$ predictions. Since a rate-coded static image delivers its total information after the $T$ timesteps passed, we choose the prediction of the last timestep as the final one.

Among the $L$ readout layers, the only predictions taken into account are from the last layer of our architecture because deep hierarchical representations have an improved representation in comparison to the lower layers.

%% file: 5_experiments.tex
\section{Proof-of-Concept Experiment}
\label{sec:experiments}

\subsection{Experimental Protocol}
\label{subsec:experimental_protocol}
We evaluate our method on the Oxford-IIIT-Pet dataset \cite{Oxford-IIIT-Pet}. It consists of 7349 RGB images. Each image represents a single cat/dog in the foreground. The dataset originally contains a foreground segmentation for each image. We convert this segmentation mask into a single bounding box surrounding the represented pet. For our experiment, we convert the images into grayscale. The training split is composed of 6000 randomly selected images and the testing split contains the remaining images. Since there is strictly one bounding box per image, we evaluate our model with the mean intersection over union metric (mIoU).

\subsection{Implementation Details}
\label{subsec:implementation_details}

We implement our SNN model by adapting the original DECOLLE \cite{DECOLLE} implementation made with PyTorch. Our experiment is performed during 100 epochs in a machine equipped with one NVIDIA 2080Ti GPU. The batch size is set to 16. We use the AdaMax optimizer \cite{kingma2014adam} with $\beta_1 = 0$, $\beta_2 = 0.95$ and a smooth L1 loss. The learning rate is initialized to $10^{-9}$. Images are resized to 176x240 pixels with data augmentation applied. Data augmentation includes random horizontal flipping and random brightness changes. As for the hyper-parameters specific to the DECOLLE method, we keep the same values as in the original paper \cite{DECOLLE}.

\subsection{Results}

\begin{figure}
    \centering
    \subfigure[Predictions]
    {
        \includegraphics[width=1.0in]{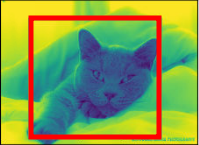}
        \includegraphics[width=1.0in]{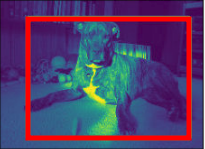}
        \includegraphics[width=1.0in]{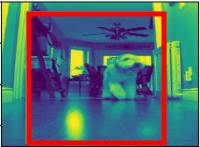}
        \label{fig:predictions}
    }
    \subfigure[Ground truths]
    {
        \includegraphics[width=1.0in]{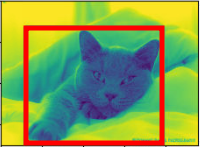}
        \includegraphics[width=1.0in]{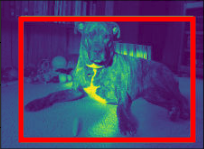}
        \includegraphics[width=1.0in]{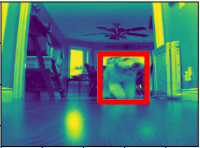}
        \label{fig:ground_truths}
    }
    \caption{Qualitative results.}
    \label{fig:qualitative}
\end{figure}

Our experiment achieves promising results of \textbf{63.2 \% mIoU} on the test set. In addition, we visually verify the consistency of the predicted bounding boxes. Fig. \ref{fig:qualitative} shows representative examples of predictions on the test set. The bounding boxes are mostly predicted with good accuracy (i.e. between 55\% and 70\% mIoU). However, rare images are still poorly predicted ($\le$ 15\% mIoU). We observe that the problem comes from images where the object is small or not in the foreground. To explain it, we have two hypotheses: \textbf{1) Data Imbalance :} our method suffers from a common data imbalance problem. Most images in Oxford-IIIT-Pet are close-up pictures of pets. Therefore, the model generalizes these examples and does not learn the rare and hard examples of pets in the background. \textbf{2) Output Conversion :} our conversion strategy (defined in Sec. \ref{subsec:output_conversion_policy}) is still not optimal and leads to a loss of information. Consequently, the predicted coordinates are not perfectly represented.

The reported results of this experiment validate the proof-of-concept and confirm the ability of state-of-the-art supervised learning rules to train SNNs in order to perform more complex vision tasks. Even if the evaluated experiment is still limited compared to state-of-the-art deep learning tasks, we argue that it can be a first step towards modern computer vision tasks. %(object detection, segmentation, ...).

%% file: 6_conclusion.tex
\section{Discussion and Conclusion}
\label{sec:conclusion}
In this work, we designed an encoder-decoder DCSNN architecture based on DECOLLE \cite{DECOLLE}, a supervised local synaptic plasticity rule, to perform object localization on a static image dataset. By evaluating our model on static grayscale natural images, we proved the ability of SNNs trained with state-of-the-art supervised algorithms to deal with modern computer vision tasks. To the best of our knowledge, this is the first work on a complex vision task using supervised SNNs and static images. Finally, we proposed an intuitive technique to obtain a final prediction from the numerous readout layers predictions. Many introduced mechanisms can be further investigated in future works, such as the optimal neural coding or the output conversion strategy. Moreover, more vision tasks can be studied using supervised SNNs. 

Even though our proposed DCSNN shows encouraging results with static images converted with rate coding, this strategy does not fully exploit the asynchronous event-based capacity of SNNs. On the other hand, event-based sensors \cite{survey-event-based} can lead to sparser and fully asynchronous vision models. Consequently, in future works, we want to further explore the performance of our architecture while using event-based cameras, which enable the treatment of spatio-temporal data instead of only static images. In addition, using event-based cameras could lead to a fully spike-based object tracking solution.

%% file: main.bbl
% Generated by IEEEtran.bst, version: 1.14 (2015/08/26)
\begin{thebibliography}{10}
\providecommand{\url}[1]{#1}
\csname url@samestyle\endcsname
\providecommand{\newblock}{\relax}
\providecommand{\bibinfo}[2]{#2}
\providecommand{\BIBentrySTDinterwordspacing}{\spaceskip=0pt\relax}
\providecommand{\BIBentryALTinterwordstretchfactor}{4}
\providecommand{\BIBentryALTinterwordspacing}{\spaceskip=\fontdimen2\font plus
\BIBentryALTinterwordstretchfactor\fontdimen3\font minus
  \fontdimen4\font\relax}
\providecommand{\BIBforeignlanguage}[2]{{%
\expandafter\ifx\csname l@#1\endcsname\relax
\typeout{** WARNING: IEEEtran.bst: No hyphenation pattern has been}%
\typeout{** loaded for the language `#1'. Using the pattern for}%
\typeout{** the default language instead.}%
\else
\language=\csname l@#1\endcsname
\fi
#2}}
\providecommand{\BIBdecl}{\relax}
\BIBdecl

\bibitem{survey-deep}
W.~Liu, Z.~Wang, X.~Liu, N.~Zeng, Y.~Liu, and F.~E. Alsaadi, ``A survey of deep
  neural network architectures and their applications,'' \emph{Neurocomputing},
  vol. 234, pp. 11--26, 2017.

\bibitem{survey_DL_compression}
Y.~Cheng, D.~Wang, P.~Zhou, and T.~Zhang, ``A survey of model compression and
  acceleration for deep neural networks,'' \emph{arXiv preprint
  arXiv:1710.09282}, 2017.

\bibitem{maass1997networks}
W.~Maass, ``Networks of spiking neurons: the third generation of neural network
  models,'' \emph{Neural networks}, vol.~10, no.~9, pp. 1659--1671, 1997.

\bibitem{hodgkin1952quantitative}
A.~L. Hodgkin and A.~F. Huxley, ``A quantitative description of membrane
  current and its application to conduction and excitation in nerve,''
  \emph{The Journal of physiology}, vol. 117, no.~4, pp. 500--544, 1952.

\bibitem{neuromorphic_STDP}
B.~Linares-Barranco, T.~Serrano-Gotarredona, L.~A. Camu{\~n}as-Mesa, J.~A.
  Perez-Carrasco, C.~Zamarre{\~n}o-Ramos, and T.~Masquelier, ``On
  spike-timing-dependent-plasticity, memristive devices, and building a
  self-learning visual cortex,'' \emph{Frontiers in neuroscience}, vol.~5,
  p.~26, 2011.

\bibitem{BP_SNN}
J.~H. Lee, T.~Delbruck, and M.~Pfeiffer, ``Training deep spiking neural
  networks using backpropagation,'' \emph{Frontiers in neuroscience}, vol.~10,
  p. 508, 2016.

\bibitem{falez2019multi}
P.~Falez, P.~Tirilly, I.~M. Bilasco, P.~Devienne, and P.~Boulet,
  ``Multi-layered spiking neural network with target timestamp threshold
  adaptation and stdp,'' in \emph{2019 International Joint Conference on Neural
  Networks (IJCNN)}.\hskip 1em plus 0.5em minus 0.4em\relax IEEE, 2019, pp.
  1--8.

\bibitem{tavanaei2019deep}
A.~Tavanaei, M.~Ghodrati, S.~R. Kheradpisheh, T.~Masquelier, and A.~Maida,
  ``Deep learning in spiking neural networks,'' \emph{Neural Networks}, vol.
  111, pp. 47--63, 2019.

\bibitem{DECOLLE}
J.~Kaiser, H.~Mostafa, and E.~Neftci, ``Synaptic plasticity dynamics for deep
  continuous local learning (decolle),'' \emph{Frontiers in Neuroscience},
  vol.~14, p. 424, 2020.

\bibitem{Oxford-IIIT-Pet}
O.~M. Parkhi, A.~Vedaldi, A.~Zisserman, and C.~V. Jawahar, ``Cats and dogs,''
  in \emph{IEEE Conference on Computer Vision and Pattern Recognition}, 2012.

\bibitem{Conversion_method}
Y.~Cao, Y.~Chen, and D.~Khosla, ``Spiking deep convolutional neural networks
  for energy-efficient object recognition,'' \emph{International Journal of
  Computer Vision}, vol. 113, no.~1, pp. 54--66, 2015.

\bibitem{conversion_sota}
A.~Sengupta, Y.~Ye, R.~Wang, C.~Liu, and K.~Roy, ``Going deeper in spiking
  neural networks: Vgg and residual architectures,'' \emph{Frontiers in
  neuroscience}, vol.~13, p.~95, 2019.

\bibitem{Spiking-YOLO}
S.~Kim, S.~Park, B.~Na, and S.~Yoon, ``Spiking-yolo: spiking neural network for
  energy-efficient object detection,'' in \emph{Proceedings of the AAAI
  Conference on Artificial Intelligence}, vol.~34, no.~07, 2020, pp.
  11\,270--11\,277.

\bibitem{STDP}
N.~Caporale and Y.~Dan, ``Spike timing--dependent plasticity: a hebbian
  learning rule,'' \emph{Annu. Rev. Neurosci.}, vol.~31, pp. 25--46, 2008.

\bibitem{falez_thesis}
P.~Falez, ``{Improving Spiking Neural Networks Trained with Spike Timing
  Dependent Plasticity for Image Recognition},'' Theses, {Universit{\'e} de
  Lille}, Oct. 2019.

\bibitem{kirkland2020spikeseg}
P.~Kirkland, G.~Di~Caterina, J.~Soraghan, and G.~Matich, ``Spikeseg: Spiking
  segmentation via stdp saliency mapping,'' in \emph{2020 International Joint
  Conference on Neural Networks (IJCNN)}.\hskip 1em plus 0.5em minus
  0.4em\relax IEEE, 2020, pp. 1--8.

\bibitem{zenke2018superspike}
F.~Zenke and S.~Ganguli, ``Superspike: Supervised learning in multilayer
  spiking neural networks,'' \emph{Neural computation}, vol.~30, no.~6, pp.
  1514--1541, 2018.

\bibitem{neftci2019surrogate}
E.~O. Neftci, H.~Mostafa, and F.~Zenke, ``Surrogate gradient learning in
  spiking neural networks: Bringing the power of gradient-based optimization to
  spiking neural networks,'' \emph{IEEE Signal Processing Magazine}, vol.~36,
  no.~6, pp. 51--63, 2019.

\bibitem{lee2020enabling}
C.~Lee, S.~S. Sarwar, P.~Panda, G.~Srinivasan, and K.~Roy, ``Enabling
  spike-based backpropagation for training deep neural network architectures,''
  \emph{Frontiers in neuroscience}, vol.~14, 2020.

\bibitem{FPN}
T.-Y. Lin, P.~Doll{\'a}r, R.~Girshick, K.~He, B.~Hariharan, and S.~Belongie,
  ``Feature pyramid networks for object detection,'' in \emph{Proceedings of
  the IEEE conference on computer vision and pattern recognition}, 2017, pp.
  2117--2125.

\bibitem{borst1999information}
A.~Borst and F.~E. Theunissen, ``Information theory and neural coding,''
  \emph{Nature neuroscience}, vol.~2, no.~11, pp. 947--957, 1999.

\bibitem{kingma2014adam}
D.~P. Kingma and J.~Ba, ``Adam: A method for stochastic optimization,''
  \emph{arXiv preprint arXiv:1412.6980}, 2014.

\bibitem{survey-event-based}
G.~Gallego, T.~Delbruck, G.~Orchard, C.~Bartolozzi, B.~Taba, A.~Censi,
  S.~Leutenegger, A.~Davison, J.~Conradt, K.~Daniilidis \emph{et~al.},
  ``Event-based vision: A survey,'' \emph{arXiv preprint arXiv:1904.08405},
  2019.

\end{thebibliography}
